\documentstyle[a4wide]{article}

\begin{document}

\author{Walter Daelemans\thanks{Visiting fellow at NIAS (Netherlands
Institute for Advanced Studies), Wassenaar, The Netherlands.} \\ CL \&
AI, Tilburg University\\ P.O.Box 90153, 5000 LE Tilburg \\ The
Netherlands \\ {\tt walter.daelemans@kub.nl}\and Peter Berck and
Steven Gillis\\ Linguistics, University of Antwerp\\
Universiteitsplein 1, 2610 Wilrijk \\ Belgium \\ {\tt
steven.gillis@uia.ua.ac.be}\\ {\tt peter.berck@uia.ua.ac.be}}

\title{
{\tt \small Proceedings of COLING 1996, Copenhagen}\\
Unsupervised Discovery of Phonological Categories through
Supervised Learning of Morphological Rules\\
}

\date{}

\maketitle

\begin{abstract}
We describe a case study in the application of {\em symbolic machine
learning} techniques for the discovery of linguistic rules and
categories.  A supervised rule induction algorithm is used to learn to
predict the correct diminutive suffix given the phonological
representation of Dutch nouns. The system produces rules which are
comparable to rules proposed by linguists. Furthermore, in the process
of learning this morphological task, the phonemes used are grouped
into phonologically relevant categories. We discuss the relevance of
our method for linguistics and language technology.
\end{abstract}

\section{Introduction}

This paper shows how machine learning techniques can be used to induce
{\em linguistically relevant} rules and categories from data.
Statistical, connectionist, and machine learning induction
(data-oriented approaches) are currently used mainly in language
engineering applications in order to alleviate the linguistic
knowledge acquisition bottleneck (the fact that lexical and
grammatical knowledge usually has to be reformulated from scratch
whenever a new application has to be built or an existing application
ported to a new domain), and to solve problems with robustness and
coverage inherent in knowledge-based (theory-oriented, hand-crafting)
approaches.  Linguistic relevance or inspectability of the induced
knowledge is usually not an issue in this type of research. In
linguistics, on the other hand, it is usually agreed that while
computer modeling is a useful (or essential) tool for enforcing
internal consistency, completeness, and empirical validity of the
linguistic theory being modeled, its role in formulating or evaluating
linguistic theories is minimal.

In this paper, we argue that machine learning techniques can also
assist in linguistic theory formation by providing a new tool for the
evaluation of linguistic hypotheses, for the extraction of rules from
corpora, and for the discovery of useful linguistic categories.  As a
case study, we apply Quinlan's C4.5 inductive machine learning method
(Quinlan, 1993) to a particular linguistic task (diminutive formation
in Dutch) and show that it can be used (i) to test linguistic
hypotheses about this process, (ii) to discover interesting
morphological rules, and (iii) discover interesting phonological
categories. Nothing hinges on our choice of C4.5 as a rule induction
mechanism. We chose it because it is an easily available and
sophisticated instance of the class of rule induction algorithms.

A second focus of this paper is the interaction between supervised and
unsupervised machine learning methods in linguistic discovery. In
supervised learning, the learner is presented a set of examples (the
experience of the system).  These examples consist of an input--output
association (in our case, e.g., a representation of a noun as input,
and the corresponding diminutive suffix as output). Unsupervised
learning methods do not provide the learner with information about the
output to be generated; only the inputs are presented to the learner
as experience, not the target outputs. 

Unsupervised learning is necessarily more limited than supervised
learning; the only information it has to construct categories is the
similarity between inputs. Unsupervised learning has been successfully
applied e.g. for the discovery of syntactic categories from corpora on
the basis of distributional information about words (Finch and Chater
1992, Hughes 1994, Sch\"{u}tze 1995).  We will show that it is
possible and useful to make use of unsupervised learning {\em
relative} to a particular task which is being learned in a supervised
way. In our experiment, phonological categories are discovered in an
unsupervised way, as a side-effect of the supervised learning of a
morphological problem. We will also show that this raises interesting
questions about the task-dependence of linguistic category systems.

\section{Supervised Rule Induction with C4.5}

For the experiments, we used C4.5 (Quinlan, 1993).  Although several
decision tree and rule induction variants have been proposed, we chose
this program because it is widely available and reasonably well
tested.  C4.5 is a TDIDT (Top Down Induction of Decision Trees)
decision tree learning algorithm which constructs a decision tree on
the basis of a set of examples (the training set). This decision tree
has tests (feature names) as nodes, and feature values as branches
between nodes. The leaf nodes are labeled with a category name and
constitute the output of the system. A decision tree constructed on
the basis of examples is used after training to assign a class to
patterns. To test whether the tree has actually learned the problem,
and has not just memorized the items it was trained on, the {\em
generalization} accuracy is measured by testing the learned tree on a
part of the dataset not used in training.

The algorithm for the construction of a C4.5 decision tree can be
easily stated.  Given are a training set $T$ (a collection of
examples), and a finite number of classes $C_{1}$ ... $C_{n}$.

\begin{enumerate}
\item 
If $T$ contains one or more cases all belonging to the same class
$C_{j}$, then the decision tree for $T$ is a leaf node with category
$C_{j}$.
\item 
If $T$ is empty, a category has to be found on the basis of other
information (e.g. domain knowledge). The heuristic used here is that
the most frequent class in the initial training set is used.
\item 
If $T$ contains different classes then 
\begin{enumerate}
\item Choose a test (feature) with a finite number
of outcomes (values), and partition $T$ into subsets of examples that
have the same outcome for the test chosen. The decision tree consists
of a root node containing the test, and a branch for each outcome,
each branch leading to a subset of the original set.
\item Apply the procedure recursively to subsets created this way. 
\end{enumerate}
\end{enumerate}

In this algorithm, it is not specified which test to choose to split a
node into subtrees at some point. Taking one at random will usually
result in large decision trees with poor generalization performance,
as uninformative tests may be chosen. Considering all possible trees
consistent with the data is computationally intractable, so a reliable
heuristic test selection method has to be found. The method used in
C4.5 is based on the concept of {\em mutual information} (or {\em
information gain}). Whenever a test has to be selected, the feature is
chosen with the highest information gain. This is the feature that
reduces the information entropy of the training (sub)set on average
most, when its value would be known. For the computation of
information gain, see Quinlan (1993).

Decision trees can be easily and automatically transformed into sets
of if-then rules (production rules), which are in general easier to
understand by domain experts (linguists in our case). In C4.5 this
tree-to-rule transformation involves additional statistical evaluation
resulting sometimes in a rule set more understandable and accurate
than the corresponding decision tree.

The C4.5 algorithm also contains a {\em value grouping} method which,
on the basis of statistical information, collapses different values
for a feature into the same category. That way, more concise decision
trees and rules can be produced (instead of several different branches
or rule conditions for each value, only one branch or condition has to
be defined, making reference to a class of values). The algorithm
works as a heuristic search of the search space of all possible
partitionings of the values of a particular feature into sets, with
the formation of homogeneous nodes (nodes representing examples with
predominantly the same category) as a heuristic guide.  See Quinlan
(1993) for more information.

\section{Diminutive Formation in Dutch}

In the remainder of this paper, we will describe a case study of using
C4.5 to test linguistic hypotheses and to discover regularities and
categories. The case study concerns allomorphy in Dutch diminutive
formation, ``one of the more vexed problems of Dutch phonology (...)
[and] one of the most spectacular phenomena of modern Dutch
morphophonemics'' (Trommelen 1983).  Diminutive formation is a
productive morphological rule in Dutch. Diminutives are formed by
attaching a form of the Germanic suffix {\em -tje} to the singular
base form of a noun. The suffix shows allomorphic variation (Table
\ref{variation}).

\begin{table}[h]
\begin{center}
\begin{tabular}{l|l|l}
Noun & Form & Suffix \\
\hline
huis (house) & huisje & {\em -je} \\
man (man) & mannetje & {\em -etje\/} \\
raam (window) & raampje & {\em -pje\/} \\
woning (house) & woninkje & {\em -kje\/} \\
baan (job) & baantje & {\em -tje\/} \\
\end{tabular}
\caption{Allomorphic variation in Dutch diminutives.}\label{variation}
\end{center}
\end{table}

The frequency distribution of the different categories is given in
Table \ref{frequency}. We distinguish between database frequency
(frequency of a suffix in a list of 3900 diminutive forms of nouns we
took from the CELEX lexical database\footnote{Developed by the Center
for Lexical Information, Nijmegen. Distributed by the Linguistic Data
Consortium.}) and corpus frequency (frequency of a suffix in the text
corpus on which the word list was based).

\begin{table}[h]
\begin{center}
\begin{tabular}{|l|r|r|r|}
\hline
Suffix & Frequency & Database \% & Corpus \% \\
\hline
tje & 1897 & 48.7\% & 50.9\% \\
je & 1462 & 37.5\% & 30.4\% \\
etje & 357 & 9.7\% & 10.9\% \\
pje & 104 & 2.7\% & 4.0\% \\
kje & 77 & 2.0\% & 3.8\% \\
\hline
\end{tabular}
\caption{Lexicon and corpus frequency of allomorphs.}\label{frequency}
\end{center}
\end{table}

Historically, different analyses of diminutive formation have taken a
different view of the rules that govern the choice of the diminutive
suffix, and of the linguistic concepts playing a role in these rules
(see e.g. Te Winkel 1866, Kruizinga 1915, Cohen 1958, and references
in Trommelen 1983). In the latter, it is argued that diminutive
formation is a local process, in which concepts such as word stress
and morphological structure (proposed in the earlier analyses) do not
play a role. The rhyme of the last syllable of the noun is necessary
and sufficient to predict the correct allomorph.  The natural
categories (or features) which are hypothesised in her rules include
{\em obstruents}, {\em sonorants}, and the class of {\em bimoraic
vowels} (consisting of long vowels, diphtongs and schwa).

Diminutive formation is a small linguistic domain for which different
competing theories have been proposed, and for which different
generalizations (in terms of rules and linguistic categories) have
been proposed. What we will show next is how machine learning
techniques may be used to (i) test competing hypotheses, (ii) discover
generalizations in the data which can then be compared to the
generalizations formulated by linguists, and (iii) discover
phonological categories in an unsupervised way by supervised learning
of diminutive suffix prediction.

\section{Experiments}

For each of the 3900 nouns we collected, the following information was
kept.
\begin{enumerate}
\item
The phoneme transcription describing the syllable structure (in terms
of onset, nucleus, and coda) of the last three syllables of the
word. Missing slots are indicated with {\tt =}.
\item
For each of these three last syllables the presence or absence of
stress. 
\item
The corresponding diminutive allomorph, abbreviated to E (-etje), T
(-tje), J (-je), K (-kje), and P (-pje). This is the `category' of the
word to be learned by the learner.
\end{enumerate}

Some examples are given below (the word itself and
its gloss are provided for convenience and were not used in the
experiments).

{\small
\begin{verbatim}
- b i = - z @ = + m A nt J biezenmand (basket)
= = = = = = = = + b I x  E big (pig)
= = = = + b K = - b a n  T bijbaan (side job)
= = = = + b K = - b @ l  T bijbel (bible)
\end{verbatim}}

\subsection{Experimental Method}

The experimental set-up used in all experiments consisted of a
ten-fold cross-validation experiment (Weiss \& Kulikowski 1991). In
this set-up, the database is partitioned ten times, each with a
different 10\% of the dataset as the test part, and the remaining 90\%
as training part. For each of the ten simulations in our experiments,
the test part was used to test generalization performance.  The
success rate of an algorithm is obtained by calculating the average
accuracy (number of test pattern categories correctly predicted) over
the ten test sets in the ten-fold cross-validation experiment.

\subsection{Learnability}

The experiments show that the diminutive formation problem is
learnable in a data-oriented way (i.e. by extraction of regularities
from examples, without any a priori knowledge about the
domain\footnote{Except syllable structure.}). The average accuracy on
unseen test data of 98.4\% should be compared to baseline performance
measures based on probability-based guessing. This baseline would be
an accuracy of about 40\% for this problem. This shows that the
problem is almost perfectly learnable by induction.  It should be
noted that CELEX contains a number of coding errors, so that some of
the `wrong' allomorphs predicted by the machine learning system were
actually correct, we did not correct for this.

In the next three sections, we will describe the results of the
experiments; first on the task of comparing conflicting theoretical
hypotheses, then on discovering linguistic generalizations, and
finally on unsupervised discovery of phonological categories.

\section{Linguistic Hypothesis Testing}

On the basis of the analysis of Dutch diminutive formation by
Trommelen (1983), discussed briefly in Section 3, the following
hypotheses (among others) can be formulated.

\begin{enumerate}
\item
Only information about the last syllable is relevant in predicting the
correct allomorph.
\item
Information about the onset of the last syllable is irrelevant in
predicting the correct allomorph.
\item
Stress is irrelevant in predicting the correct allomorph.
\end{enumerate}

In other words, information about the rhyme of the last syllable of a
noun is necessary and sufficient to predict the correct allomorph of
the diminutive suffix. To test these hypotheses, we performed four
experiments, training and testing the C4.5 machine learning algorithm
with four different corpora. These corpora contained the following
information.
\begin{enumerate}
\item
All information (stress, onset, nucleus, coda) about the three last
syllables (3-SYLL corpus).
\item
All information about the last syllable (SONC corpus).
\item
Information about the last syllable without stress (ONC corpus).
\item
Information about the last syllable without stress and onset (NC
corpus).
\end{enumerate}

\subsection{Results}

Table \ref{results} lists the learnability results. The generalization
error is given for each allomorph for the four different training
corpora.

\begin{table}[h]
{\footnotesize
\begin{center}
\begin{tabular}{|l||rr|rr|rr|rr|}
\hline
& \multicolumn{8}{|c|}{Errors and Error percentages} \\
\hline
\hline
Suffix & \multicolumn{2}{|c|}{\sc 3 syll} & \multicolumn{2}{|c|}{{\sc sonc}} & 
\multicolumn{2}{|c|}{{\sc onc}} & \multicolumn{2}{|c|}{{\sc nc}}  \\
\hline
{\it Total} & 61 & 1.6 & 79 & 2.0 & 80 & 2.0 & 77 & 2.0 \\
\hline
{\it -tje\/} & 13 & 0.7 & 13 & 0.7 & 14 & 0.7 & 14 & 0.7 \\
{\it -je\/} & 16 & 1.1 & 15 & 1.0 & 16 & 1.1 & 14 & 1.0 \\
{\it -etje\/} & 26 & 7.3 & 49 & 13.7 & 48 & 13.5 & 44 & 12.3 \\
{\it -kje\/} & 4 & 5.2 & 0 & 0 & 0 & 0 & 0 & 0 \\
{\it -pje\/} & 2 & 1.9 & 2 & 1.9 & 2 & 1.9 & 5 & 4.8  \\
\hline
\end{tabular}
\caption{Error of C4.5 on the different corpora.}\label{results}
\end{center}}
\end{table}

The overall best results are achieved with the most elaborate corpus
(containing all information about the three last syllables),
suggesting that, contra Trommelen, important information is lost by
restricting attention to only the last syllable. As far as the
different encodings of the last syllable are concerned, however, the
learnability experiment coroborates Trommelen's claim that stress and
onset are not necessary to predict the correct diminutive allomorph.
When we look at the error rates for individual allomorphs, a more
complex picture emerges. The error rate on {\em -etje} dramatically
increases (from 7\% to 14\%) when restricting information to the last
syllable. The {\em -kje} allomorph, on the other hand, is learned
perfectly on the basis of the last syllable alone. What has happened
here is that the learning method has {\em overgeneralized} a rule
predicting -kje after the velar nasal, because the data do not contain
enough information to correctly handle the notoriously difficult
opposition between words like {\em leerling} (pupil, takes {\em
-etje}) and {\em koning} (king, takes {\em -kje}).  Furthermore, the
error rate on {\em -pje} is doubled when onset information is left out
from the corpus.

We can conclude from these experiments that although the broad lines
of the analysis by Trommelen (1983) are correct, the learnability
results point at a number of problems with it (notably with {\em -kje}
versus {\em -etje} and with {\em -pje}). We will move now to the use
of inductive learning algorithms as a generator of generalizations
about the domain, and compare these generalizations to the analysis of
Trommelen.

\section{Supervised Learning of Linguistic Generalizations}

When looking only at the rhyme of the last syllable (the NC corpus),
the decision tree generated by C4.5 looks as follows:

{\footnotesize
\begin{verbatim}
Decision Tree:

coda in {rk,nt,lt,rt,p,k,t,st,s,ts,rs,rp,f,
         x,lk,Nk,mp,xt,rst,ns,nst,
         rx,kt,ft,lf,mt,lp,ks,ls,kst,lx}: J
coda in {n,=,l,j,r,m,N,rn,rm,w,lm}:
|   nucleus in {I,A,},O,E}:
|   |   coda in {n,l,r,m}: E
|   |   coda in {=,j,rn}: T
|   |   coda in {rm,lm}: P
|   |   coda = N:
|   |   |   nucleus = I: K
|   |   |   nucleus in {A,O,E}: E
|   nucleus in {K,a,e,u,M,@,y,o,i,L,),|,<}:
|   |   coda in {n,=,l,j,r,rn,w}: T
|   |   coda = m: P
\end{verbatim}
}

Notice that the phoneme representation used by CELEX (called DISC) is
shown here instead of the more standard IPA font, and that the value
grouping mechanism of C4.5 has created a number of phonological
categories by collapsing different phonemes into sets indicated by
curly brackets.

This decision tree should be read as follows: first check the coda (of
the last syllable). If it ends in an obstruent, the allomorph is {\em
-je}.  If not, check the nucleus. If it is bimoraic, and the coda is
/m/, decide {\em -pje}, if the coda is not /m/, decide {\em -tje}.
When the coda is not an obstruent, the nucleus is short and the coda
is /ng/, we have to look at the nucleus again to decide between {\em
-kje} and {\em -etje} (this is where the overgeneralization to {\em
-kje} for words in -ing occurs). Finally, the coda (nasa-liquid or
not) helps us distinguish between {\em -etje} and {\em -pje} for those
cases where the nucleus is short. It should be clear that this tree
can easily be formulated as a set of rules without loss of accuracy.

An interesting problem is that the {\em -etje} versus {\em -kje}
problem for words ending in {\em -ing} could not be solved by
referring only to the last syllable (C4.5 and any other statistically
based induction algorithm overgeneralize to {\em -kje}). The following
is the knowledge derived by C4.5 from the full corpus, with all
information about the three last syllables (the 3 SYLL corpus). We
provide the rule version of the inferred knowledge this time.

{\footnotesize
\begin{verbatim}
        Default class is -tje

1.      IF coda last is /lm/ or /rm/
        THEN  -pje

2.      IF nucleus last is [+bimoraic]
           coda last is  /m/
        THEN  -pje

3.      IF coda last is /N/
        THEN IF  nucleus penultimate is empty
                 (monosyllabic word) or schwa 
             THEN  -etje
             ELSE  -kje 

4.      IF nucleus last is [+short]
           coda last is [+nas] or [+liq]
        THEN  -etje
 
5.      IF coda last is [+obstruent]
        THEN  -je
\end{verbatim}}

The default class is {\em -tje}, which is the allomorph chosen when
none of the other rules apply. This explains why this rule set looks
simpler than the decision tree earlier.

The first thing which is interesting in this rule set, is that only
three of the twelve presented features (coda and nucleus of the last
syllable, nucleus of the penultimate syllable) are used in the
rules. Contrary to the hypothesis of Trommelen, apart from the rhyme
of the last syllable, the nucleus of the penultimate syllable is taken
to be relevant as well.

The induced rules roughly correspond to the previous decision tree,
but in addition a solution is provided to the {\em -etje} versus {\em
-kje} problem for words ending in {\em -ing} (rule 3) making use of
information about the nucleus of the penultimate syllable. Rule 3
states that words ending in /ng/ get {\em -etje} as diminutive
allomorph when they are monosyllables (nucleus of the penultimate
syllable is empty) or when they have a schwa as penultimate nucleus,
and {\em -kje} otherwise. As far as we now, this generalization has
not been proposed in this form in the published literature on
diminutive formation.

We conclude from this part of the experiment that the machine learning
method has succeeded in extracting a sophisticated set of linguistic
rules from the examples in a purely data-oriented way, and that these
rules are formulated at a level that makes their use in the
development of linguistic theories possible.

\section{Discovery of Phonological Categories}

To structure the phoneme inventory of a language, linguists define
features. These can be interpreted as sets of speech sounds
(categories): e.g. the category (or feature) $labial$ groups those
speech sounds that involve the lips as an active articulator.  Speech
sounds belong to different categories, i.e., are defined by different
features. E.g. $t$ is voiceless, a coronal, and a stop.  Categories
proposed in phonology are inspired by articulatory, acoustic or
perceptual phonetic differences between speech sounds. They are also
proposed to allow an optimally concise or elegant formulation of rules
for the description of phonological or morphological processes. E.g.,
the so-called major class features (obstruents, nasals, liquids,
glides, vowels) efficiently explain syllable structure computation,
but are of little use in the definition of rules describing
assimilation. For assimilation, place of articulation features are
best used. This situation has led to the proposal of many different
phonological category systems. 

While constructing the decision tree (see previous section), several
phonologically relevant categories are `discovered' by the value
grouping mechanism in C4.5, including the nasals, the liquids, the
obstruents, the short vowels, and the bimoraic vowels.  This last
category corresponds completely with the (then new) category
hypothesised by Trommelen and containing the long vowels, the
diphtongs and the schwa.  In other words, the learning algorithm has
discovered this set of phonemes to be a useful category in solving the
diminutive formation problem by providing an extensional definition of
it (a list of the instances of the category).

This raises the question of the task-dependence of linguistic
categories. Similar experiments in Dutch plural formation, for
example, fail to produce the category of bimoraic vowels, and for some
tasks, categories show up which have no ontological status in
linguistics. In other words, making category formation dependent on
the task to be learned, undermines the traditional linguistic ideas
about absolute, task-independent (and even language-independent)
categories. We present here a new methodology with which this
fundamental issue in linguistics can be investigated: category systems
extracted for different tasks in different languages can be studied to
see which categories (if any) truely have a universal status. This is
subject for further research. It would also be useful to study the
induced categories when intensional descriptions (feature
representations) are used as input instead of extensional descriptions
(phonemes).

We also experimented with a simpler alternative to the computationally
complex heuristic category formation algorithm used by C4.5. This
method is inspired by machine learning work on value difference
metrics (Stanfill \& Waltz, 1986; Cost \& Salzberg, 1993). Starting
from the training set of the supervised learning experiment (the set
of input--output mappings used by the system to extract rules), we
select a particular feature (e.g. the coda of the last syllable), and
compute a table associating with each possible value of the feature
the number of times the pattern in which it occurs was assigned to
each different category (in this case, each of the the five
allomorphs). This produces a table with for each value a distribution
over categories. This table is then used in standard clustering
approaches to derive categories of values (in this case
consonants). The following is one of these clustering results. The
example shows that this computationally simple approach also succeeds
in discovering categories in an unsupervised way on the basis of data
for supervised learning.

{\footnotesize
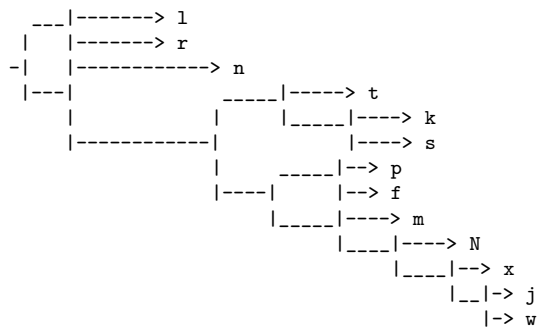
\begin{verbatim}
  ___|-------> l
 |   |-------> r
-|   |------------> n
 |---|             _____|-----> t
     |            |     |_____|----> k
     |------------|           |----> s
                  |     _____|--> p
                  |----|     |--> f
                       |_____|----> m
                             |____|----> N
                                  |____|--> x
                                       |__|-> j
                                          |-> w
\end{verbatim}
}

Several categories, relevant for diminutive formation, such as
liquids, nasals, the velar nasal, semi-vowels, fricatives etc., are
reflected in this hierarchical clustering.

\section{Conclusion}

We have shown by example that machine learning techniques can
profitably be used in linguistics as a tool for the comparison of
linguistic theories and hypotheses or for the discovery of new
linguistic theories in the form of linguistic rules or categories.

The case study we presented concerns diminutive formation in Dutch,
for which we showed that (i) machine learning techniques can be used to
corroborate and falsify some of the existing theories about the
phenomenon, and (ii) machine learning techniques can be used to
(re)discover interesting linguistic rules (e.g. the rule solving the
{\em -etje} versus {\em -kje} problem) and categories (e.g. the
category of bimoraic vowels).

The extracted system can of course also be used in language technology
as a data-oriented system for solving particular linguistic tasks (in
this case diminutive formation). In order to test the usability of the
approach for this application, we compared the performance of the
extracted rule system to the performance of the hand-crafted rule
system proposed by Trommelen.  Table \ref{trommelen} shows for each
allomorph the number of errors by the C4.5 rules (trained using corpus
NC, i.e. only the rhyme of the last syllable) as opposed to an
implementation of the rules suggested by Trommelen. One problem with
the latter is that they often suggest more than one allomorph (the
rules are not mutually exclusive). In those cases where more than one
rule applies, a choice was made at random.

\begin{table}[h]
\begin{center}
\begin{tabular}{|l|r|r|}
\hline
Suffix & Trommelen & C4.5\\
\hline
{\em -tje} & 53 & 11\\
{\em -je} & 12 & 12\\
{\em -etje} & 28 & 39\\
{\em -kje} & 38 & 0\\
{\em -pje} & 21 & 4\\
\hline
Total & 152 & 66\\
\hline
\end{tabular}
\caption{Comparison of accuracy between handcrafted and induced
rules.}\label{trommelen}
\end{center}
\end{table}

The comparison shows that C4.5 did a good job of finding an elegant
and accurate rule-based description of the problem. This rule set is
useful both in linguistics (for evaluation, refinement, and discovery
of theories) and in language technology.


\begin{thebibliography}{mmmmmmmm 1966}

\bibitem[Cohen 1958]{Cohen:1958} Cohen, A.
Het Nederlandse diminutiefsuffix; een morfologische proeve.  {\em De
Nieuwe Taalgids}, 51, 40-45, 1958.

\bibitem[Cost 1993]{Cost:1993}
Cost, S. and Salzberg, S.  `A weighted nearest neighbor algorithm for
learning with symbolic features.' {\em Machine Learning}, 10, 57--78,
1993.

\bibitem[Finch 1992]{Finch:1992}
Finch, S. \& N. Chater. `Bootstrapping Syntactic Categories Using
Statistical Methods', in: W.Daelemans \& D.Powers (eds.), {\em
Background and Experiments in Machine Learning of Natural Language},
Tilburg University, ITK, 1992.

\bibitem[Hughes 1994]{Hughes:1994}
Hughes, J. `Automatically Acquiring a Classification of Words', PhD
dissertation, University of Leeds, School of Computer Studies, 1994

\bibitem[Krui\-sin\-ga 1915]{Kruisinga:1915} Kruisinga, E.
De vorm van verkleinwoorden.  {\em De Nieuwe Taalgids}, 9, 96-97,
1915.

\bibitem[Quinlan 1993]{Quinlan:1993} Quinlan, J. R.
{\em C4.5 Programs for machine learning} 1993.

\bibitem[Schuetze 1995]{Schuetze:1995}
Sch\"utze, H., Ambiguity in Language Learning: Computational and
Cognitive Models, PhD dissertation, Stanford University, Department of
Linguistics, 1995.

\bibitem[Stanfill 1986]{Stanfill:1986}
Stanfill, C. and Waltz, D.L. `Toward Memory-based Reasoning'.  {\it
Communications of the ACM} 29, 1986, 1213--1228.

\bibitem[Trommelen 1983]{Trommelen:1983} Trommelen, M.T.G.
{\em The syllable in Dutch, with special reference to diminutive
formation.}  Foris, Dordrecht, 1983.

\bibitem[Weiss 1991]{wei-kul:91}
Weiss, S. and Kulikowski, C. (1991).
\newblock {\em Computer systems that learn}.
\newblock Morgan Kaufmann, San Mateo.

\bibitem[Te Winkel 1862]{Winkel:1866} Winkel, L.A. Te.
{\em Over de verkleinwoorden.}  De Taalgids 4: 81-116.
\end{thebibliography}
\end{document}